\documentclass[10pt,twocolumn,letterpaper]{article}

\usepackage{iccv}
\usepackage{placeins}
\usepackage{float}

\definecolor{iccvblue}{rgb}{0.21,0.49,0.74}
\usepackage[pagebackref,breaklinks,colorlinks,allcolors=iccvblue]{hyperref}

\def\paperID{*****} 
\def\confName{ICCV}
\def\confYear{2025}

\title{COMBAT: Conditional World Models for Behavioral Agent Training}

\author{
Anmol Agarwal$^{1,2}$\thanks{Equal contribution.} \quad
Sumer Singh$^{2 *}$ \quad
Pranay Meshram$^{2 *}$ \quad
Saurav Suman$^{2 *}$ \\
Andrew Lapp$^{1}$ \quad
Shahbuland Matiana$^{1}$ \quad
Louis Castricato$^{1}$ \quad
Spencer Frazier$^{1}$\thanks{Corresponding author.}  \\
$^{1}$Overworld AI \\ 
$^{2}$Indian Institute of Science Education and Research Bhopal 
}

\begin{document}
\maketitle

\begin{abstract}
Recent advances in generative AI have spurred the development of world models capable of simulating 3D-consistent environments and interactions with static objects. A significant limitation of these models is the ability to model dynamic, reactive agents which can intelligently influence and interact with the world. We introduce COMBAT, a real-time, action-controlled world model trained on the complex 1v1 fighting game Tekken 3 to address these shortcomings. Our work demonstrates that diffusion models can successfully simulate a dynamic opponent that reacts to player actions while learning its behavior implicitly.

Our approach utilizes a 1.2 billion parameter Diffusion Transformer, conditioned on latent representations from a deep compression autoencoder. We employ state-of-the-art techniques, including causal distillation and diffusion forcing to achieve real-time inference. Crucially, we observe the emergence of sophisticated agent behavior by training the model solely on single-player inputs, without any explicit supervision for the opponent's policy. Unlike traditional imitation learning methods which require complete action labels, COMBAT learns effectively from partially observed data to generate responsive behaviors for a controllable primary player (Player 1). We present our results from an extensive study and introduce novel evaluation methods to benchmark this emergent agent behavior. In the process, establishing a strong foundation for training interactive agents within diffusion-based world models.
\end{abstract}

\begin{figure*}[t]
  \centering
  \includegraphics[width=0.68\linewidth]{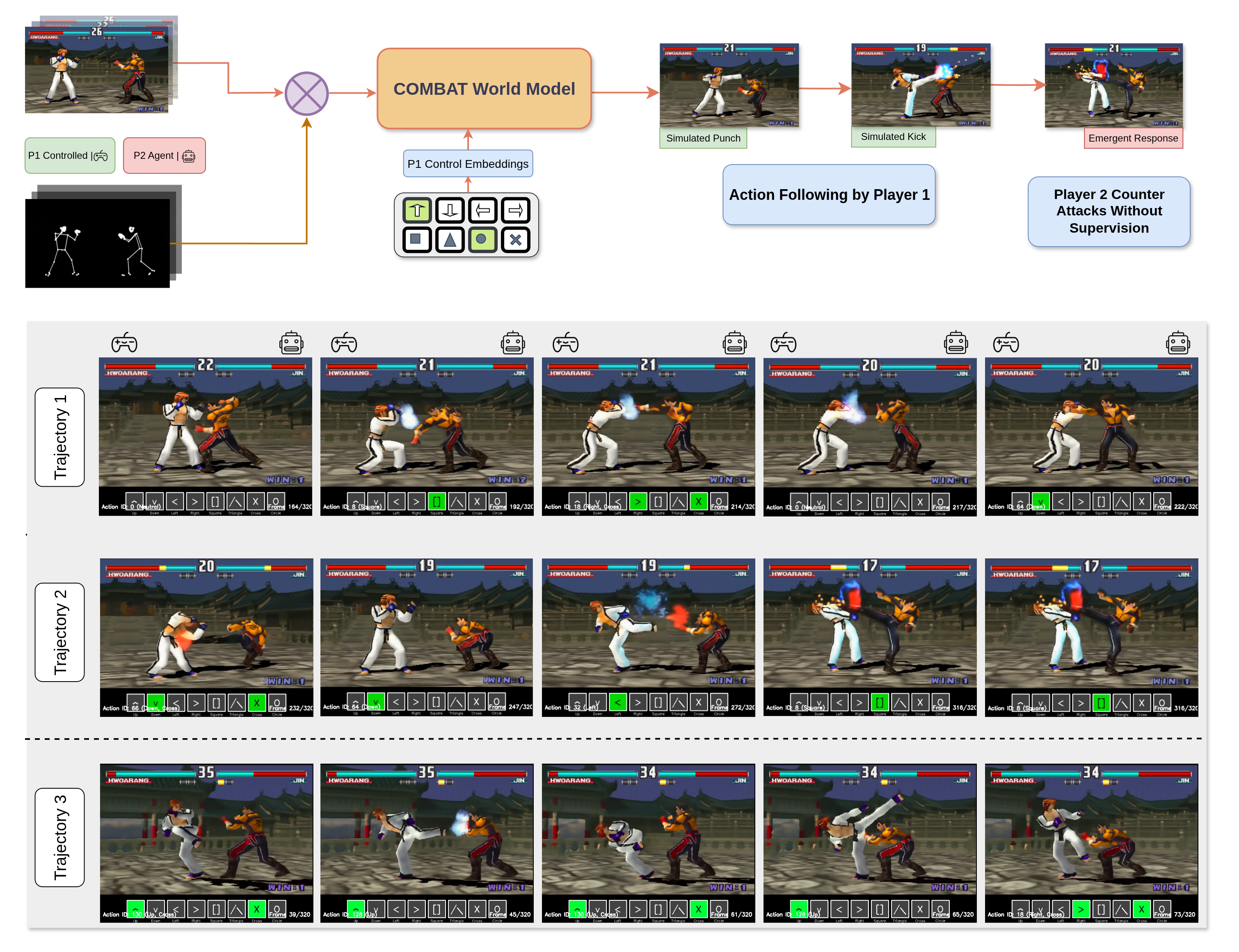}
  \caption{An overview of the COMBAT world model. (Top) The model is conditioned on the current state (visual frames and poses) and Player 1's control inputs to autoregressively predict subsequent frames. (Bottom) Three distinct generated trajectories showcase the model's ability to produce plausible, strategic counter-attacks from Player 2 as an emergent response to Player 1's actions, without direct supervision of the opponent's policy.}
  \label{fig:short}
\end{figure*}

\section{Introduction}
\label{sec:intro}

As the fidelity of video generation methods improve with increased understanding of real-world phenomena and context, interactive world models trained on gameplay and real-world data have emerged to accelerate these advances \cite{sora2024, Bruce2024GenieGI, valevsky2025}. Generating spatially and temporally consistent world simulations are the primary focus. Yet, in real-world scenarios, the most unpredictable components are reactive agents that can observe, plan, and influence their environment. This is especially evident in autonomous driving, navigation, and combat scenarios.

Recent work demonstrates that autoregressive diffusion models are very effective for world simulation. Recent advances make these models real-time through distribution matching distillation (DMD) \cite{yin2024improved, yin2025causvid} and diffusion forcing \cite{huang2025selfforcing} to overcome autoregressive drift. This work has enabled neural game simulations for first-person games such as Minecraft and CS:GO \cite{hygamecraft} and showcase excellent causal understanding of actions and their effects on generated frames.

However, real-world and game environments also contain rich information about how agents (e.g. humans, NPCs and autonomous systems) respond to environmental dynamics. Current methods could greatly benefit from learning agent behavior from this observational data, but the partial observability and unstructured nature poses significant challenges. For example, while we might observe a pedestrian changing pathing to avoid a vehicle; the exact observations and decision processes of the human agent remain hidden.

We present COMBAT (\textbf{C}onditional world \textbf{M}odel for \textbf{B}ehavioral \textbf{A}gent \textbf{T}raining), an interactive world model that learns underlying agent behavior and movement dynamics directly from partially observed multi-agent systems. By training a world model on Tekken 3 gameplay with conditioning only on Player 1's input, we observe emergent tactical behavior in Player 2 without explicit behavioral supervision. We select Tekken 3 as it provides an ideal controlled environment with clear visual feedback, deterministic game mechanics, diverse movesets, and frame-precise timing requirements.

Our approach uses a 1.2B parameter diffusion transformer trained on 1.2M frames across 1,000 gameplay rounds. We first train a Deep Compression AutoEncoder (DCAE) \cite{dcae} to obtain highly compressed latent representations, then train the world model to generate temporally consistent gameplay sequences. COMBAT successfully learns to control Player 1 from conditioning signals, while Player 2 emerges with realistic combat behaviors including blocking, counterattacking, and combo execution. Through decoder distillation and CausVid DMD \cite{yin2025causvid} techniques, we achieve real-time generation at interactive frame rates.

We introduce novel benchmarking methods to evaluate emergent agent behavior. This includes measurement of behavioral diversity and tactical understanding. Our extensive analysis demonstrates that world models can serve as a new paradigm for learning agent behaviors from observational data, with implications for multi-agent AI systems beyond gaming.

\section{Related Work}
\label{sec:related_work}

\subsection{Video Diffusion Models}
The remarkable success of diffusion models in image synthesis \cite{dalle2021, sd2022} has naturally inspired their extension to video generation. Early approaches adapted U-Net architectures from image models, achieving results in short-form video synthesis \cite{svd2023, guo2024animatediff}. However, the convolutional nature of U-Net presents challenges for video: it struggles to capture long-range temporal dependencies and scales poorly with sequence length, often leading to temporal incoherence. Our work is positioned at the intersection of generative world models, video diffusion architectures, and behavioral modeling. We review key advancements in these areas to contextualize our contribution.

To address these limitations, Transformer-based video models have emerged. Peebles et al. \cite{Peebles2022DiT} demonstrates that Diffusion Transformers (DiT) could surpass U-Nets with respect to image generation with superior scaling properties. Subsequent work has applied this architecture to video. Models such as W.A.L.T \cite{gupta2023walt} and CogVideoX \cite{yang2025cogvideox} show that DiT self-attention mechanisms effectively model complex spatiotemporal relationships in video data, enabling longer, more coherent sequences. Our work builds on this foundation, employing a DiT backbone tailored for action-conditioned dynamics in interactive environments.

\subsection{Neural Game Engines and World Models}
Recent advances demonstrate that generative models can serve as complete, neural game engines, replacing traditional rendering and state update logic. As an example, GameGAN learns to imitate 2D games from raw pixels and actions using GANs with explicit memory modules \cite{Kim2020_GameGan}. More recently, diffusion transformers have become dominant for this task.

GameNGen is another example of a fully neural DOOM engine that generates frames conditioned on past frames and actions enabling real-time simulation \cite{valevsky2025}. DIAMOND trains diffusion-based world models achieving state-of-the-art RL performance while producing playable Counter-Strike simulations \cite{diamond2024}. GameGen-X extends this, training on million-clip datasets to enable long-horizon, interactive open-world gameplay \cite{che2025gamegenx}.

These methods validate that neural models can learn complex game dynamics from observational data. Our work adopts similar architectural foundations but introduces a novel objective: modeling emergent behavior of uncontrolled opponents that arises solely from conditioning on controllable player actions.

\subsection{Multi-Modal and Behavioral World Models}
While traditional world models focus on visual prediction, recent work has enabled greater fidelity and behavioral learning. Our work adopts joint RGB-pose representation to enforce structural consistency in character movements.

In parallel, learning agent behavior within world models has predominantly followed two paths. The first is \textbf{model-based reinforcement learning}, where an agent's policy is trained using a learned dynamics model and an extrinsic reward signal. Works like \textbf{DreamerV3} exemplify this, achieving mastery in diverse domains by learning behaviors entirely within the latent space of a world model \cite{hafner2024masteringdiversedomainsworld}. The second path is \textbf{imitation learning}, which learns policies from expert demonstrations. Methods like \textbf{Generative Adversarial Imitation Learning (GAIL)} require explicit state-action supervision for all agents to mimic expert behavior \cite{GANimitation}.

Our approach diverges from both paradigms. We demonstrate that complex, reactive multi-agent behaviors emerge implicitly as a property of world modeling itself, without engineered reward signals and using only partially observed data where just one agent's actions are provided as a condition.

\subsection{Optimization Techniques for Interactive Generation}
Real-time interactive generation requires addressing both architectural efficiency and sampling speed. Recent advances in attention mechanisms include FlexAttention \cite{dong2024flexattention}, which enables flexible attention patterns, and Longformer \cite{Beltagy2020Longformer}, which combines local sliding-window attention with global context. We incorporate local-global attention patterns inspired by these works to balance efficiency with temporal coverage.

For sampling efficiency, Distribution Matching Distillation (DMD) \cite{yin2024improved, yin2025causvid} and diffusion forcing \cite{huang2025selfforcing} have proven effective techniques for reducing sampling steps while mitigating autoregressive drift. These techniques enable real-time neural simulation for complex games \cite{Bruce2024GenieGI, valevsky2025}. We adapt DMD through CausVid distillation to achieve interactive frame rates while preserving behavioral quality.

The Muon optimizer \cite{jordan2024muon} introduces orthogonalization into momentum-based updates, improving conditioning of weight updates and outperforming AdamW in training speed benchmarks. We incorporate Muon optimization to enhance training efficiency of our large-scale diffusion transformers.
\section{Method}

Our proposed and studied approach, \textbf{COMBAT}, learns to simulate a complex, multi-agent environment by training a generative world model on video observations. World models have shown promise in mastering diverse domains~\cite{hafner2024masteringdiversedomainsworld} and creating interactive environments~\cite{Bruce2024GenieGI, yang2023learningrealworldsim}. We extend this paradigm to a competitive fighting game, where the model must learn the opponent's behavior without explicit action labels.

\subsection{Problem Formulation}

The task is as follows: Primarily, learning a conditional video generation model that implicitly captures an opponent's policy. We select the fighting game \emph{Tekken 3} as our environment for three key reasons:
\begin{enumerate}
    \item \textbf{Bounded Temporal Dependency}: The game state is largely Markovian, where 
    \[
    P(s_{t+1}\mid s_{\leq t}) \approx P(s_{t+1}\mid s_{t-k:t}),
    \]
    for a small history window $k$, since all relevant information is contained within recent frames.
    \item \textbf{Rich Action Space}: Characters possess diverse movesets, with over 40 unique actions and complex combos, providing a challenging domain for behavior modeling.
    \item \textbf{Strategic Depth}: Success requires a blend of rapid reactions and long-term tactical planning.
\end{enumerate}

\textbf{Formal Problem Statement:}  
Given a dataset of partially observed multi-agent trajectories
\[
D = \{(s_t, a_t^{(1)}, s_{t+1})\}_{t=1}^T,
\]
where $s_t \in \mathbb{R}^{H \times W \times 3}$ is a game frame and $a_t^{(1)} \in \{0,1\}^8$ is the observed multi-hot input for Player 1. The actions of Player 2, $a_t^{(2)}$, remain unobserved. Our objective is to learn a conditional world model
\[
P_\theta(s_{t+1}\mid s_{t-k:t}, a_{t-k:t}^{(1)})
\]
that can accurately predict subsequent frames.

\textbf{Key Innovation:}  
Unlike traditional imitation learning methods that require explicit action supervision for all agents~\cite{GANimitation}, COMBAT is trained without Player 2's action labels. The model must infer Player 2's policy,
\[
\pi^{(2)}(a_t^{(2)} \mid s_t, a_t^{(1)}),
\]
as an emergent property of generating temporally consistent and plausible multi-agent interactions. This forces the world model to learn reactive and strategic opponent behavior implicitly.

\begin{figure*}[t]
    \centering
    \begin{subfigure}[b]{0.6\textwidth} 
        \centering
        \includegraphics[width=\linewidth]{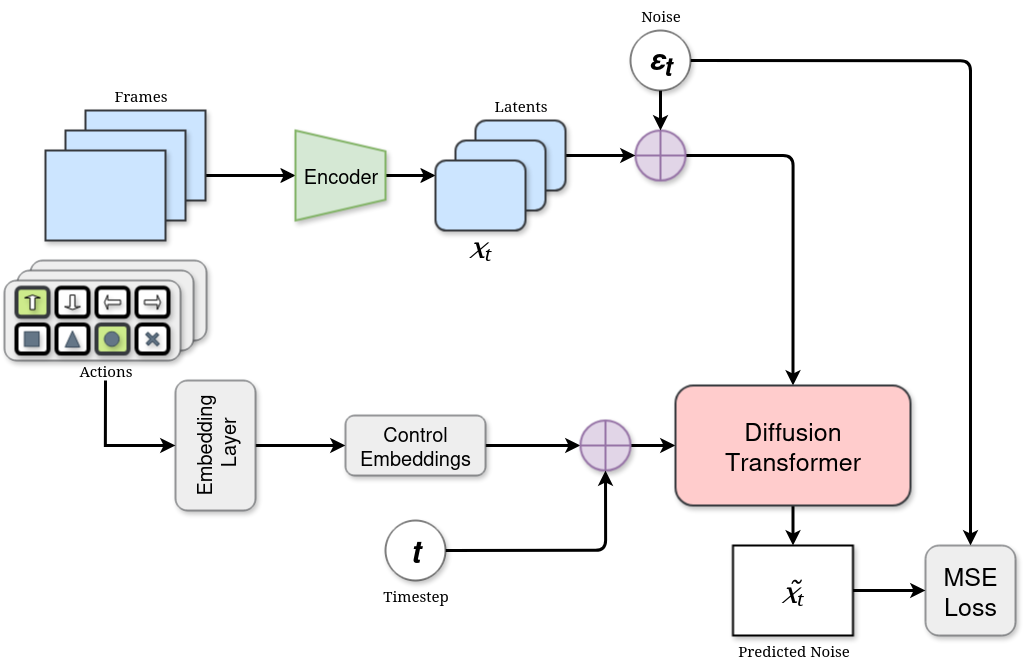}
        \caption{Training overview of COMBAT World Model.}
        \label{fig:first_sub}
    \end{subfigure}
    \hfill 
    \begin{subfigure}[b]{0.3\textwidth} 
        \centering
        \includegraphics[width=\linewidth]{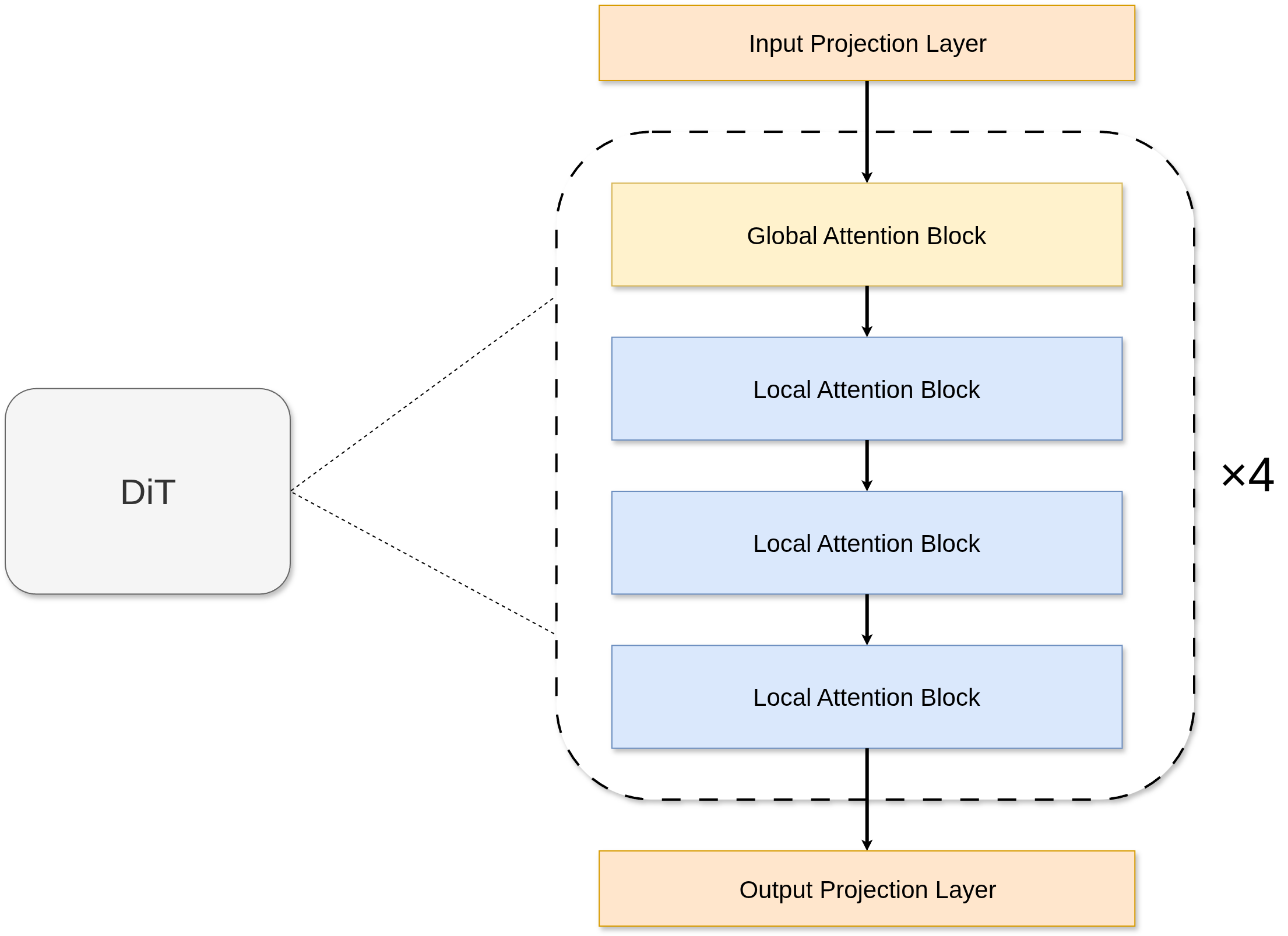}
        \caption{Every 4th DiT block has a global attention layer to capture long form context.}
        \label{fig:second_sub}
    \end{subfigure}
    \caption{Architectural diagram of the COMBAT model. (a) The end-to-end training process, where a Diffusion Transformer is conditioned on action and timestep embeddings to denoise latent frame representations. (b) The internal structure of the DiT backbone, which employs a hybrid local-global attention pattern to efficiently model long-term dependencies.}
    \label{fig:combat_results}
\end{figure*}

\subsection{Tekken 3 Gameplay Dataset}

To train our model, we collect a large-scale dataset of \emph{Tekken 3} gameplay, totaling 1{,}000 rounds (approximately 7 hours or 1.2 million frames). The dataset features a variety of characters and a balanced win--loss ratio between the two players. For each frame, captured at a resolution of $3 \times 448 \times 736$, we provide synchronized annotations including: a) action inputs for both players, b) health and timer status, c) 68-point body pose coordinates, and d) player segmentation masks. Our data collection and annotation pipeline will be made publicly available in conjunction with the publication of this paper.

\subsection{Model Architecture}

Our world model architecture integrates three main components:
\begin{itemize}
    \item a multi-modal variational autoencoder for high-ratio state compression,
    \item an embedding module for player actions and diffusion timesteps, and 
    \item a Diffusion Transformer (DiT) backbone for autoregressive prediction in the latent space.  
\end{itemize}

We train two versions of the model: one using only RGB latents and another using a joint visual--pose latent representation.

\subsubsection{Multi-Modal Latent Encoding}

To create an efficient latent representation, we first train a 340M-parameter joint RGB--pose variational autoencoder. This model learns a shared embedding space by compressing concatenated visual frames ($3 \times 448 \times 736$) and pose keypoints into a compact latent tensor of shape $128 \times 23 \times 11$. Our design is inspired by recent work in high-compression autoencoders for diffusion models~\cite{chen2025deep}. To optimize for real-time performance, the 340M-parameter decoder is subsequently distilled to a 44M-parameter version by reducing its upsampling block count, which maintains high reconstruction quality at a fraction of the computational cost.

Player 1's action history is projected into a dense embedding, encoded as a multi-hot vector over 8 buttons. This action embedding is summed with a sinusoidal time embedding for the current diffusion step, $t_{\text{emb}}$, to form the final conditioning vector for the DiT backbone.

\subsubsection{Diffusion Transformer Backbone}

The core of our generative model is a 1.2B-parameter Diffusion Transformer (DiT)~\cite{Peebles2022DiT}, which learns to denoise and predict future latent frames. The architecture consists of 16 transformer blocks with a model dimension $d_{\text{model}} = 2048$ and 16 attention heads. The conditioning vector is injected into each block via an Adaptive Layer Normalization Zero (AdaLNZero) layer, and tokenization is performed using linear projection layers for spatio-temporal rasterization, bypassing conventional patch-based embeddings.

Each DiT block executes the following sequence:  
\begin{align*}
    \text{AdaLN} &\rightarrow \text{Attention} \rightarrow \text{Gated Residual} \rightarrow \\
    &\quad \text{AdaLN} \rightarrow \text{MLP} \rightarrow \text{Gated Residual}
\end{align*}

To maintain computational tractability over long 128-frame sequences, we employ a hybrid attention strategy. Most layers use a frame-causal attention mask with a local sliding window of 16 frames, while every fourth layer applies global attention across the entire 128-frame context. This structure balances long-range dependency modeling with computational efficiency. We apply Rotary Position Embeddings (RoPE) \cite{su2023rope} across both spatial and temporal axes and utilize FlexAttention for an efficient block-sparse masking implementation.

\subsection{Accelerated Inference for Real-Time Generation}

Enabling real-time interaction is critical for gaming applications, but the iterative sampling process of diffusion models is computationally intensive. To overcome this, we significantly accelerate inference using two key optimizations.

First, we distill the fully trained model into a few-step sampler using Distribution Matching Distillation (DMD)~\cite{yin2024improved}. We adopt the CausVid DMD framework~\cite{yin2025causvid} to produce a 4-step distilled model that preserves high generative fidelity while drastically reducing inference time.  

Second, we further enhance speed by implementing static key-value caching, which reuses previously computed attention states across generation steps. These optimizations are applied to both the RGB and visual--pose world models.

\section{Experiments}
\label{sec:experiments}

To validate our claim that a conditional world model can learn reactive agent behavior from partial observations, we conduct a series of experiments on the Tekken 3 dataset. We first detail our multi-stage training pipeline and model architectures. We then introduce our evaluation benchmarks and present results comparing our primary models and their distilled variants.

\subsection{Implementation Details}
\label{sec:implementation}

Our training process is divided into three main stages: autoencoder training, world model training, and distillation for real-time inference. All models were trained on a cluster of $8 \times$ NVIDIA H200 GPUs.

\textbf{Stage 1: Autoencoder Training.}
We first train a 340M parameter Deep Compression AutoEncoder (DCAE) to learn a compact latent representation of the game environment. The autoencoder is trained for 68,000 steps (approx. 75 hours) on our 1.2 million frame Tekken dataset. It compresses raw frames ($3 \times 448 \times 736$) into a latent space of $23 \times 11$ with 128 channels. The training objective is a combination of L2 reconstruction loss, perceptual similarity loss, and a KL divergence term to regularize the latent space. For our pose-augmented model, we use an identical architecture and training setup.

\textbf{Stage 2: World Model Training.}
We train a 1.2B parameter autoregressive Diffusion Transformer (DiT) to function as the world model. The DiT architecture consists of 16 layers, 16 attention heads, and a model dimension of $d_{model} = 2048$. It employs a combination of local (16 frames) and global (128 frames) attention windows to capture both short-term and long-term temporal dependencies. The model is trained on video clips with a sequence length of 128 frames to predict the next latent frame conditioned on Player 1's actions. We train two distinct world models: one using latents from the RGB-only VAE and another using latents from the pose-augmented VAE.

\textbf{Stage 3: Distillation for Real-Time Inference.}
To achieve interactive frame rates, we employ two separate distillation techniques:
\begin{itemize}
    \item \textbf{Decoder Distillation:} We first create a lightweight VAE decoder for real-time rendering. Using student-teacher distillation, we reduce the number of upsampling blocks per stage in the decoder from four to one. This process reduces the decoder's parameter count from 340M to a nimble 44M.
    \item \textbf{Step Distillation:} We use CausVid, a Distribution Matching Distillation (DMD) method, to drastically reduce the number of required inference steps for the world model. We distill the fully-trained DiT into a 4-step variant. This distillation process converges in 2,500 steps, utilizing a combination of a DMD loss and a critic loss. We apply this technique to both the RGB-only and the pose-augmented world models.
\end{itemize}

\subsection{Evaluation Metrics and Benchmarks}
\label{benchmarks}

Evaluating emergent agent behavior presents a fundamental challenge: How do we measure intelligence that was never explicitly supervised? Traditional video metrics assess visual fidelity, while RL metrics assume access to ground-truth actions or rewards. Since COMBAT learns behavioral patterns implicitly through world modeling, we need novel evaluation approaches capable of detecting tactical competence from generated gameplay alone.

\subsubsection{Standard Perceptual Metrics}
To assess the perceptual quality of our generated trajectories, we employ a suite of standard metrics. Our evaluation protocol involves conditioning the models on real Player 1 action sequences extracted from a test set of 300 ground-truth videos (roughly 1-2 seconds in length) consisting of mixed difficulty gameplay. The generated video is then compared directly against its corresponding ground-truth counterpart from which the actions were sourced. This setup provides a stringent test of the model's ability to render deterministic outcomes based on specific actions.

We report the Fréchet Video Distance (FVD)\cite{fvd} to measure temporal coherence, the Fréchet Inception Distance (FID)\cite{fid} for per-frame visual fidelity, and LPIPS to quantify perceptual similarity. Given the high-fidelity nature of the Tekken 3 environment, achieving strong performance on these metrics against the ground truth is a robust indicator of the model's precision and world-modeling capabilities.
\begin{table}[h!]
\centering
\caption{
     All metrics are calculated on a held-out test set of 300 video clips each with 32 frames. Lower is better for all scores.
}
\label{tab:perceptual_metrics}
\begin{tabular}{@{}lccc@{}}
\toprule
\textbf{Model} & \textbf{FID} $\downarrow$ & \textbf{FVD} $\downarrow$ & \textbf{LPIPS} $\downarrow$ \\
\midrule
COMBAT: Pose & 49.7 & 593.4 & 0.05 \\
COMBAT: Non-Pose & 80.9 & 1156.6 & 0.07 \\
\midrule
\bottomrule
\end{tabular}
\end{table}


\begin{figure}[H]
    \centering
    \begin{subfigure}[b]{0.48\linewidth}
        \includegraphics[width=\linewidth]{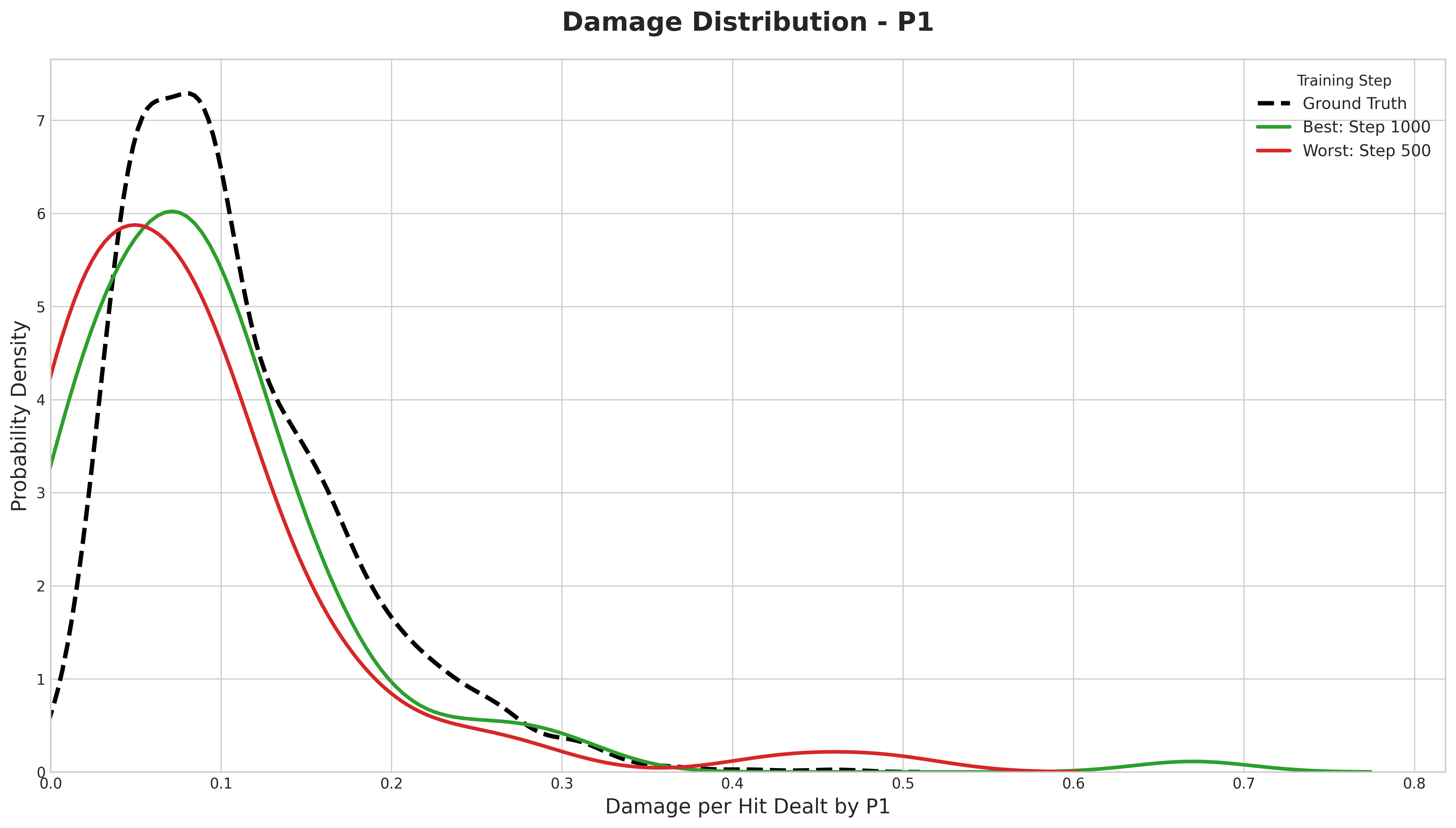}
        \caption{Player 1 Damage Distribution}
        \label{fig:p1_damage}
    \end{subfigure}
    \hfill 
    \begin{subfigure}[b]{0.48\linewidth}
        \includegraphics[width=\linewidth]{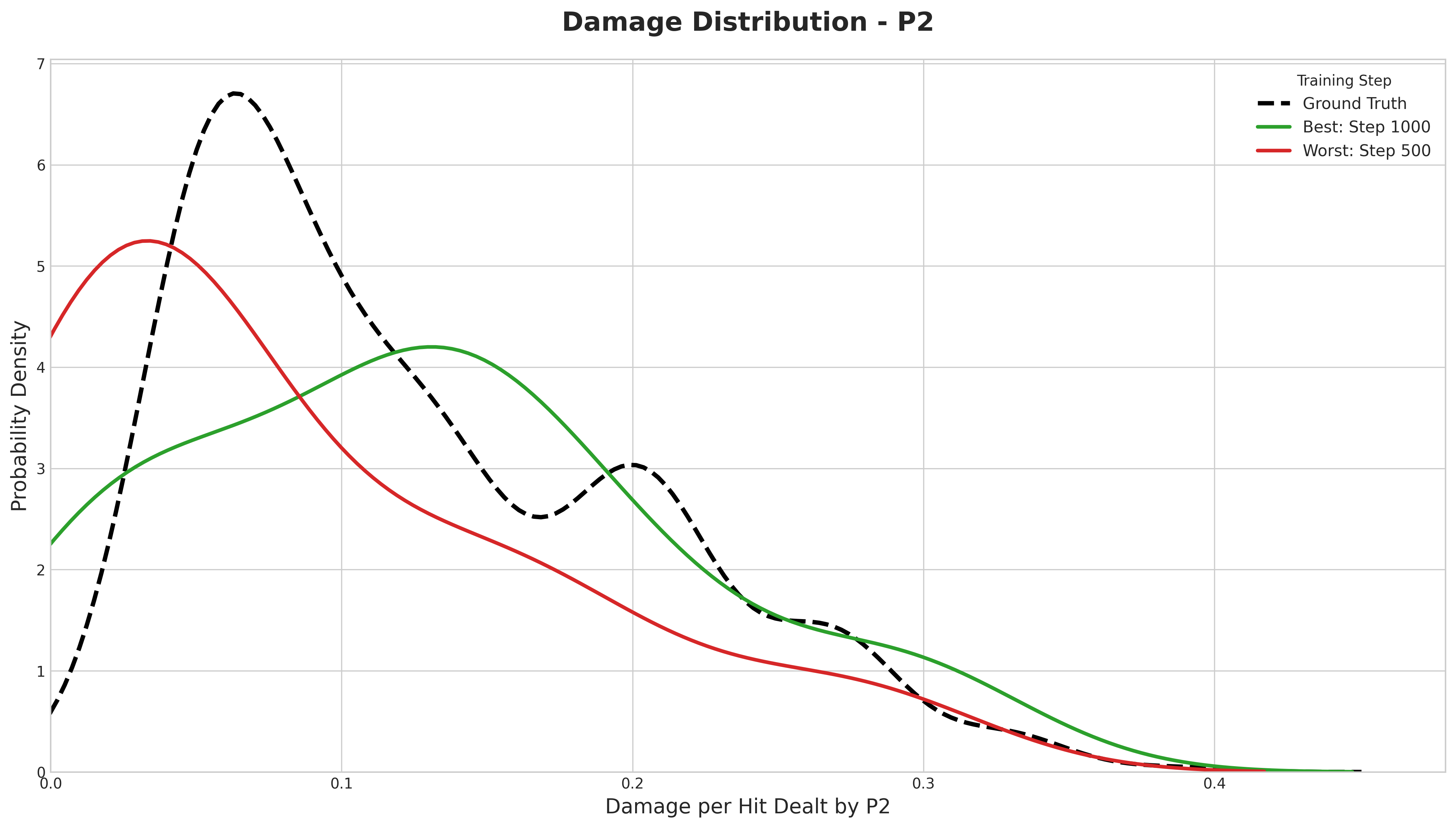}
        \caption{Player 2 Damage Distribution}
        \label{fig:p2_damage}
    \end{subfigure}

    \vspace{0.5cm} 

    \begin{subfigure}[b]{0.48\linewidth}
        \includegraphics[width=\linewidth]{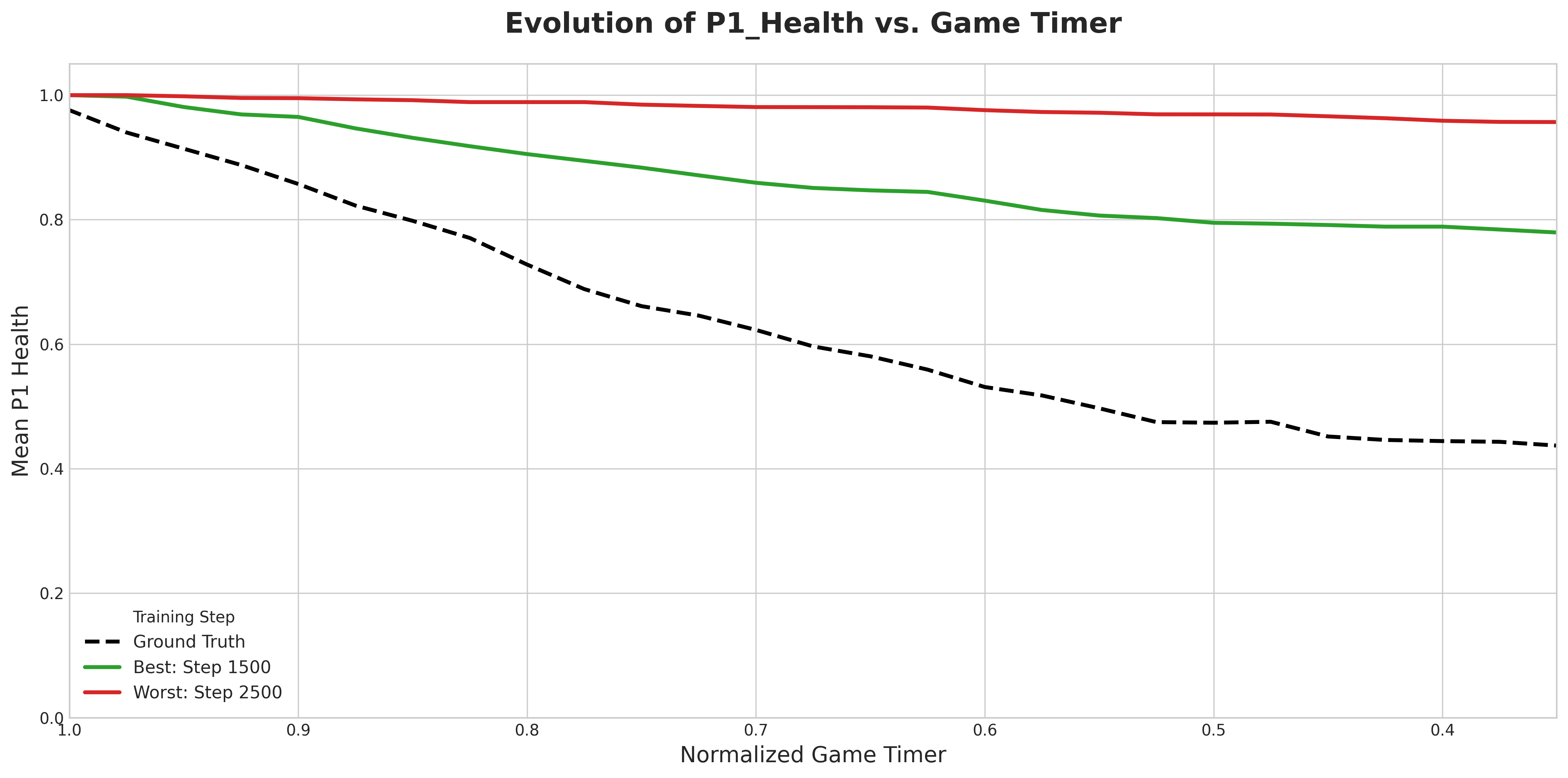}
        \caption{Player 1 Mean Health Trajectory}
        \label{fig:p1_health}
    \end{subfigure}
    \hfill 
    \begin{subfigure}[b]{0.48\linewidth}
        \includegraphics[width=\linewidth]{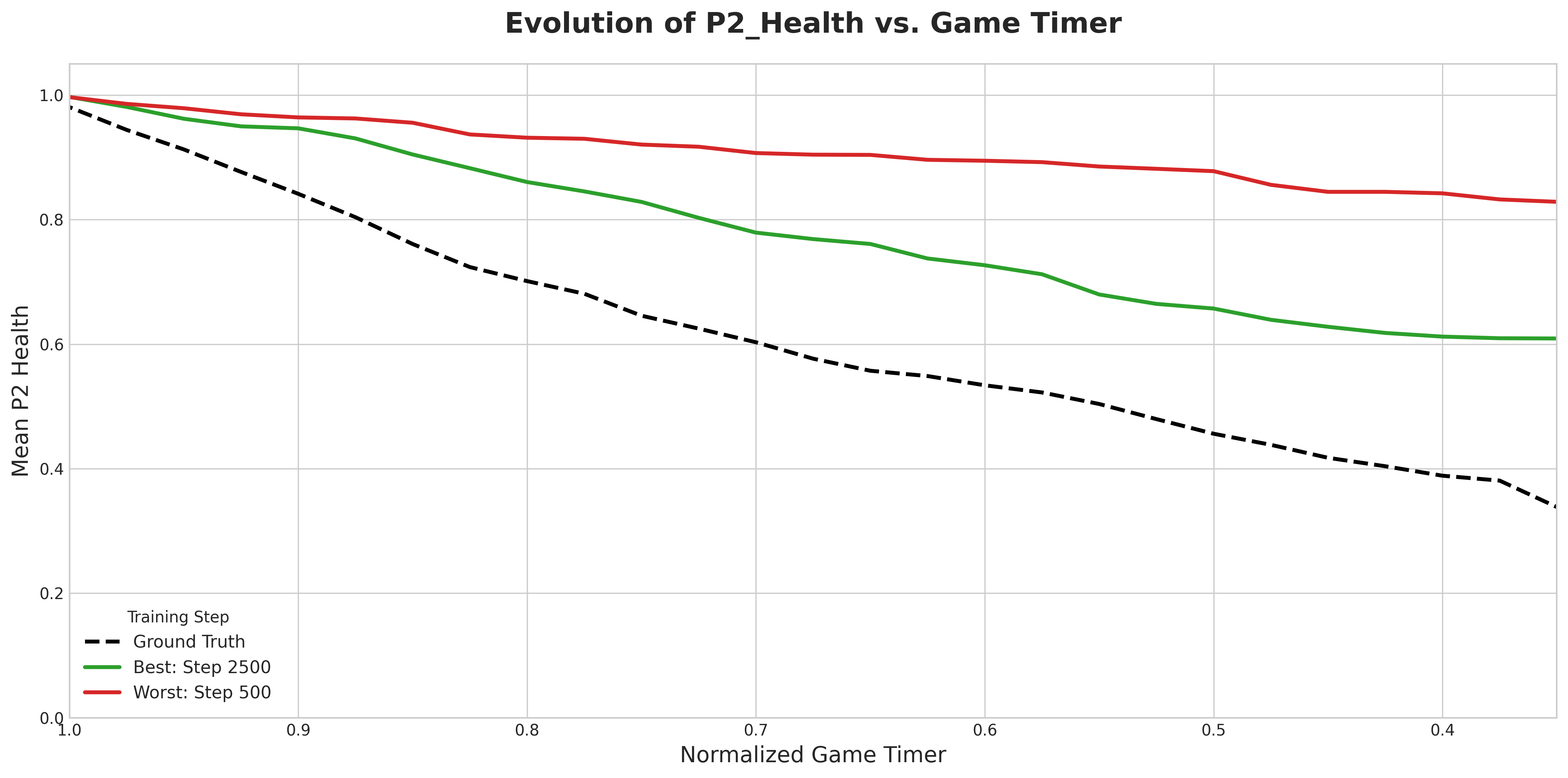}
        \caption{Player 2 Mean Health Trajectory}
        \label{fig:p2_health}
    \end{subfigure}
    
    \caption{
        \textbf{Behavioral Consistency Metrics.} A comparison of generated gameplay (COMBAT) against the ground truth. 
        \textbf{(a, b)} The per-frame damage distributions for Player 1 and Player 2, showing that our model learns a realistic mapping of actions to consequences.
        \textbf{(c, d)} The mean health trajectories over the course of a round, indicating that COMBAT captures the natural pacing of a match.
    }
    \label{fig:behavioral_metrics_grid}
\end{figure}

\subsubsection{Behavioral Consistency Metrics}
To verify that our model learns the game's intrinsic rules and pacing, we propose two metrics based on in-game health data:

\begin{itemize}
    \item \textbf{Damage Distribution Analysis:}
    This metric assesses whether the consequence of individual actions is realistic. Let $H^{(t)}_i$ denote the health of player $i \in \{1,2\}$ at frame $t$, and define per-frame damage as
    $\Delta H^{(t)}_i = \max(0, H^{(t-1)}_i - H^{(t)}_i)$. We normalize by the maximum health $H^\text{max}_i$ to obtain
    $\delta^{(t)}_i = \Delta H^{(t)}_i / H^\text{max}_i$.

    The complete distribution of damage values from all generated sequences, $\{\delta^{(t)}_{i,\text{gen}}\}$, is then compared to the distribution from all ground-truth sequences, $\{\delta^{(t)}_{i,\text{real}}\}$, using the Wasserstein distance. A lower distance signifies that the model has learned a more accurate mapping from actions to their in-game consequences.

    \item \textbf{Health Trajectory Analysis:}
    This metric evaluates the overall temporal flow of the match. Define the normalized time $s = t/T$, where $T$ is the total round duration, and let $\bar{H}^{(s)} = \frac{1}{2}\sum_i H^{(t)}_i / H^\text{max}_i$ be the average normalized health at time $s$ for a single round.

    To establish a baseline for typical match progression, we compute the \textbf{mean health trajectory} by averaging $\bar{H}^{(s)}$ across all rounds in our ground-truth test set. We do the same for our generated rounds. The similarity between these two mean trajectories is then measured using the Mean Squared Error (MSE). A lower MSE indicates that the generated gameplay, on average, exhibits a more realistic match pace.
\end{itemize}

\subsection{Human Evaluation of Emergent Behavior}

To assess the emergent behavior of Player 2, we conduct human evaluation based on observable action patterns in gameplay. Since Player 2 is trained without explicit supervision, emergent behavior is defined as actions that react naturally to Player 1's inputs, demonstrating plausible combat strategies such as timely punches, kicks, and defensive maneuvers.

We introduce two human-interpretable metrics: \textbf{Total Action Adherence (TAA)} and \textbf{Action Ratio Consistency (ARC)}. These metrics are based on human annotations of offensive actions observed in both ground-truth and generated gameplay sequences.

\subsubsection{Total Action Adherence (TAA)}

TAA measures whether the agent produces a comparable overall volume of offensive actions relative to human gameplay:

\[
\text{TAA} = \frac{G_{\text{kicks}} + G_{\text{punch}}}{O_{\text{kicks}} + O_{\text{punch}}}
\]

where $G_{\cdot}$ denotes actions performed by the generated agent, and $O_{\cdot}$ the actions performed in original gameplay.

A score of 1.0 indicates perfect adherence in activity level. Scores $>1.0$ suggest hyperactive behavior, while scores $<1.0$ indicate passive behavior.

\begin{figure}[H]
    \centering
    \includegraphics[width=0.9\linewidth]{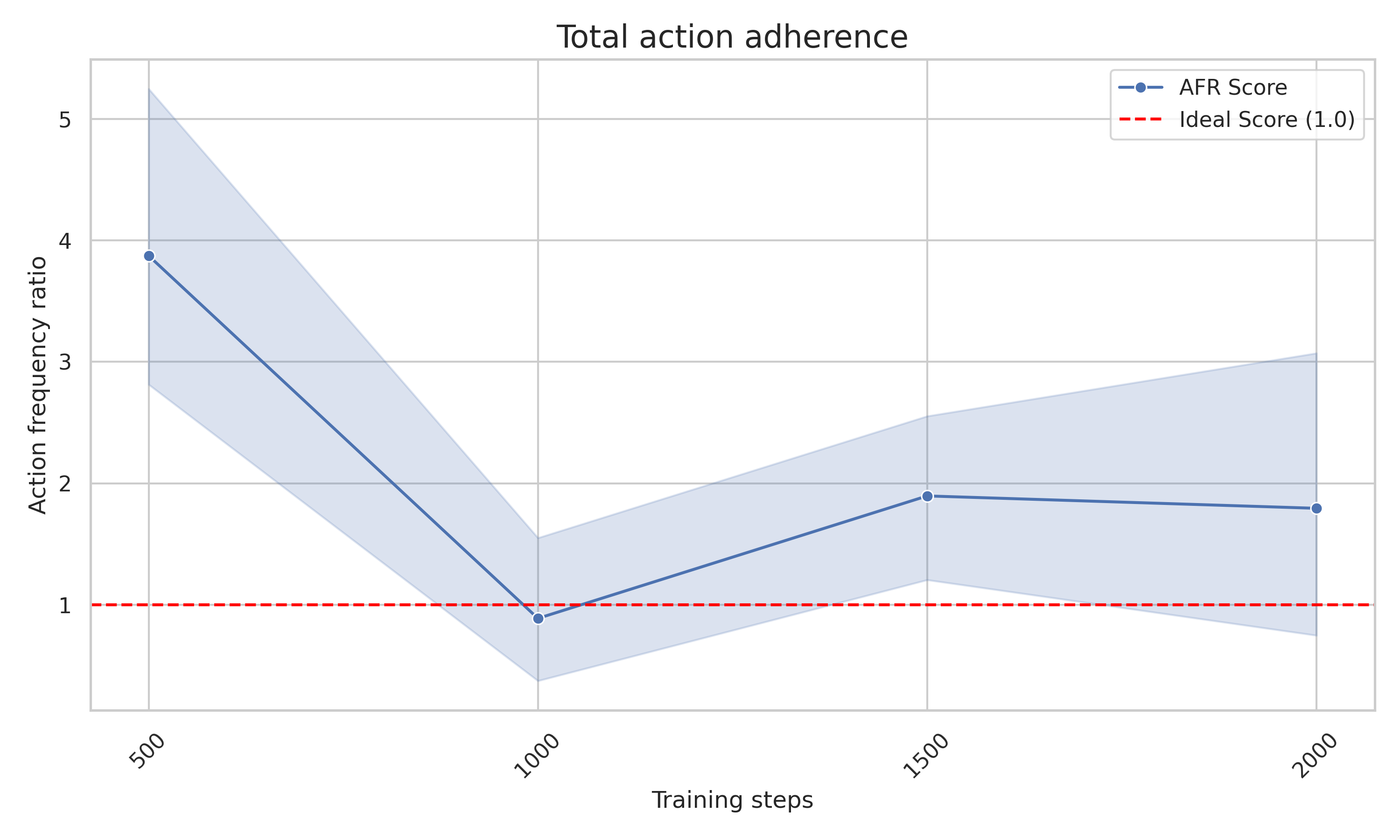}
    \caption{Total Action Adherence across training checkpoints}
    \label{fig:total_action_adherence}
\end{figure}

\subsubsection{Action Ratio Consistency (ARC)}

ARC evaluates whether the stylistic balance between punches and kicks aligns with the human player:

\[
\text{ARC} = \frac{\tfrac{G_{\text{punch}}}{G_{\text{kicks}}}}{\tfrac{O_{\text{punch}}}{O_{\text{kicks}}}}
\]

A score of 1.0 indicates identical punch-to-kick ratio as original gameplay. Scores above 1.0 reflect stronger preference for punches, while scores below 1.0 suggest heavier reliance on kicks.

\begin{figure}[H]
    \centering
    \includegraphics[width=0.9\linewidth]{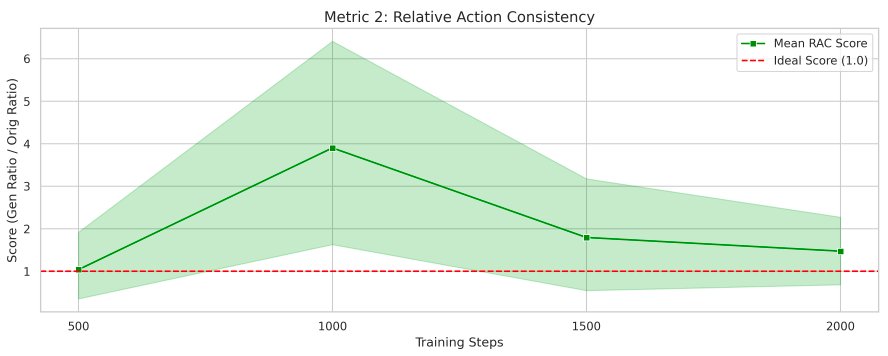}
    \caption{Action Ratio Consistency across training checkpoints}
    \label{fig:action_ratio_consistency}
\end{figure}

\subsubsection{Results}

We evaluated sequences at multiple training checkpoints. Table~\ref{tab:taa-arc} summarizes the results:

\begin{table}[h]
\centering
\begin{tabular}{lcc}
\toprule
\textbf{Training Step} & \textbf{TAA} & \textbf{ARC} \\
\midrule
Ground Truth & 1.00 & 1.00 \\
Step 500     & 3.87 & 1.04 \\
Step 1000    & 0.88 & 3.90 \\
Step 1500    & 1.90 & 1.79 \\
Step 2000    & 1.79 & 1.47 \\
\bottomrule
\end{tabular}
\caption{TAA and ARC scores at different training checkpoints compared against human gameplay.}
\label{tab:taa-arc}
\end{table}

Our evaluation shows that COMBAT successfully learns emergent Player 2 behavior through distinct phases. Initially, the model is hyperactive, generating nearly four times the offensive actions of human players (TAA = 3.87), though its punch-to-kick ratio is well-aligned (ARC = 1.04). As training progresses, the model reduces hyperactivity in further steps. Beyond step 2000, performance declines, with later checkpoints showing reduced adherence to original gameplay.

By the final training stages, the model converges toward stable, human-like combat patterns. It learns to regulate activity frequency (TAA 1.8) while achieving balanced fighting style (ARC  1.5). However, overall consistency degrades noticeably.

The pose-augmented COMBAT model significantly outperforms the RGB-only variant across visual quality metrics, confirming that explicit pose information improves generation quality.
\begin{itemize}
    
     \item\textbf{Impact of Distillation:} Our 4-step distilled models, created using CausVid DMD, retain substantial visual quality while achieving 12.5× speedup. The pose-augmented 4-step model still outperforms the full RGB-only model, demonstrating efficient distillation with minimal quality trade-off.

Qualitatively, we observe intelligent behaviors including combo execution, spatial awareness, and adaptation to Player 1's patterns. These tactical responses emerge naturally from our training process without explicit behavioral supervision.
\end{itemize}

\section{Conclusion}

In this work, we introduce COMBAT, a conditional world model that learns complex, emergent agent behavior from partially observed gameplay. Our key finding is that by conditioning the model solely on Player 1's actions it successfully learns a reactive, tactically coherent policy for Player 2 without any direct supervision. The model correctly associates the control inputs with the intended agent and generates plausible counter-attacks. This demonstrates that intricate behaviors can arise implicitly from the objective of temporal consistency.

We provide an extensive analysis of emergent behavior in world models to enable further analysis and research. We also release our large-scale \textbf{Tekken 3 dataset} complete with synchronized pose and segmentation annotations, and \textbf{open-source our pipelines} for data collection and model training.

Our approach is practical for interactive entertainment applications. Through distillation, the COMBAT world model achieves \textbf{real-time performance, operating at 85~FPS on a single NVIDIA A100 GPU}. This work represents a contribution as to how generative world models can learn implicit agent policies, and we hope it inspires further research into multi-agent behavioral modeling in complex, interactive environments.

\section{Future Work}

We identify two primary directions for future research. First, while DMD step distillation accelerates inference, it degrades agent responsiveness and attack frequency. Future work should develop distillation techniques that preserve behavioral fidelity by incorporating metrics like Action Ratio Consistency (ARC) into the optimization objective.

Second, integrating reinforcement learning (RL) finetuning could guide the world model toward goal-oriented behaviors like maximizing win-rate. This involves training a policy \textit{within} the generative model's latent space, establishing a new paradigm for intelligent agents in simulated environments.

{
    \small
    \bibliographystyle{ieeenat_fullname}
    \bibliography{main}
}


\end{document}